%% file: main.tex
\documentclass[letterpaper]{article} 
\usepackage[preprint]{aaai2027}  
\nocopyright  
\usepackage[hyphens]{url}  
\usepackage{graphicx} 
\usepackage{natbib}  
\usepackage{caption} 
\usepackage{booktabs}
\usepackage{multirow}
\usepackage{makecell}
\usepackage{tabularx}

\usepackage{amsmath}
\usepackage{amssymb}

\usepackage{enumitem}

\ifx\else
\fi

\input{preamble}

\title{When Does Sharing Work? Scaling a Shared SigLIP2 Across Image, Video, Audio, and Audio-Visual Representations}

\author{Xiaofei Sun, Peng Wang, Huayao Liu, Xiaoxiao Xu, Jun Song, Deqing Li, Lin Qu}
\affiliations{Alibaba Group}

\begin{document}
\maketitle

\input{sec/0_abstract}
\input{sec/1_intro}

\input{sec/2_related_work}

\input{sec/3_method}
\input{sec/4_data_recipe}

\input{sec/5_experiments}

\input{sec/6_conclusion}


\bibliography{main}

\end{document}

%% file: preamble.tex
\usepackage{tikz}
\usetikzlibrary{positioning, arrows.meta, fit, backgrounds, calc}



%% file: sec/0_abstract.tex
\begin{abstract}
Contrastive pretraining has advanced image-text understanding, yet image, video, and audio encoders are still trained independently, with cross-modal alignment realized mainly through post-hoc binding to a frozen anchor space.
We present \textbf{Omni-Encoder}, a general multimodal encoder that processes image, video, audio, and audio-visual inputs within a \emph{single} SigLIP-2 Vision Transformer with fully shared parameters.
Unlike anchor-centric binding or per-modality adapters, all modalities are optimized jointly end-to-end, and the framework is organized from input to training.
A modality-native input stage first maps each signal into the shared embedding space, including a 1D-convolutional audio tokenizer that converts mel-spectrograms into tokens aligned with visual patches.
A multimodal rotary position embedding parameterized by absolute timestamps then places visual frames and audio segments in a unified spatiotemporal coordinate system.
Finally, alternating-modality training interleaves image, video, audio, and temporally aligned audio-visual batches and accumulates their gradients into one update, which we find central to stable joint optimization.
Omni-Encoder achieves competitive results on 16 benchmarks across all four modalities with one shared parameter set, and joint training improves each modality over its separately trained counterpart.
\end{abstract}

%% file: sec/1_intro.tex
\section{Introduction}
\label{sec:intro}

\input{sec/Figures/fig_radar_tikz}


Contrastive pretraining has established dual-encoder vision-language models~\cite{radford2021learning,zhai2023sigmoid,tschannen2025siglip2,sun2023evaclip} as a dominant paradigm for transferable representation learning, and the same recipe has since been replicated independently for video~\cite{xu2021videoclip,wang2022internvideo,zhao2024videoprism,luo2022clip4clip,li2023umt} and audio~\cite{wu2023clap,elizalde2023clap,mei2024wavcaps}.
Human perception, in contrast, is inherently multisensory: the brain integrates what is seen, heard, and felt into a unified representation through joint, time-locked exposure across the senses~\cite{ghazanfar2006multisensory,calvert2001crossmodal,bavelier2002cross,stein2008multisensory}.
This contrast raises a natural question for representation learning---whether image, video, audio, and audio-visual streams can be served by \emph{one} encoder with a single set of shared parameters, in the way a single brain serves all senses.


Existing efforts toward such unification remain partial, and can be summarized into two paradigms.
The first trains a separate encoder per modality and aligns them post-hoc, typically by binding non-anchor modalities to a frozen anchor space~\cite{guzhov2022audioclip,girdhar2023imagebind,zhu2024languagebind}.
The second shares backbone layers across modalities while retaining modality-specific components such as adapters or experts~\cite{bao2024onepeace,wang2023beit3}.
Across both paradigms, three limitations recur:
(i)~\emph{redundant capacity}, since overlapping concepts are relearned through disjoint parameters;
(ii)~\emph{constrained joint optimization}, since frozen anchors or private adapter stacks limit how far non-anchor modalities can be optimized together with the rest;
and (iii)~\emph{weak native cross-modal correspondence}, since temporal correspondence between audio and video is rarely modeled inside the encoder itself~\cite{arandjelovic2017look,morgado2021avid,gong2023cavmae}.
As a result, a single encoder that processes all four modality groups through the same parameters---without degrading any of them---remains under-explored.


We present \textbf{Omni-Encoder}, a general omni-modal encoder that processes image, video, audio, and audio-visual inputs within a \emph{single} SigLIP-2 Vision Transformer with fully shared parameters, organized along the path from input to training.
A modality-native input stage first maps each signal into the shared embedding space, including a 1D-convolutional audio tokenizer that converts mel-spectrograms into tokens aligned with visual patches.
A multimodal rotary position embedding parameterized by absolute timestamps then places visual frames and audio segments in a unified spatiotemporal coordinate system, so that audio-visual co-occurrence becomes a positional fact readable by the same self-attention.
Finally, an alternating-modality training strategy interleaves image, video, audio, and temporally aligned audio-visual batches and accumulates their gradients into one balanced update, which we find central to stable joint optimization.


We evaluate Omni-Encoder on 16 benchmarks spanning image classification, image-text retrieval, video-text retrieval, audio-text retrieval, and audio-visual retrieval, under matched parameter scale and evaluation protocol, so that comparisons reflect design choices rather than model size or test-set selection.
With one shared parameter set, Omni-Encoder achieves competitive results across all four modality groups (Figure~\ref{fig:radar}), and joint training tends to improve each modality over its separately trained counterpart, suggesting the potential for positive cross-modal transfer rather than interference.
We hope these findings provide a practical reference for unified multimodal representation learning.


\begin{figure*}[t]
\centering
\definecolor{cImg}{RGB}{52,120,200}
\definecolor{cVid}{RGB}{230,140,40}
\definecolor{cAud}{RGB}{60,160,100}
\definecolor{cVA}{RGB}{140,90,180}
\definecolor{cLoss}{RGB}{198,55,55}
\definecolor{cEnc}{RGB}{58,75,135}
\resizebox{\textwidth}{!}{%
\begin{tikzpicture}[
  font=\scriptsize,
  every node/.style={align=center, inner sep=1.5pt},
  mbox/.style={rounded corners=2pt, draw=#1, fill=#1!12, very thick,
               minimum width=3.6cm, minimum height=1.3cm},
  lbl/.style={font=\tiny, text=black!70},
  pill/.style={rounded corners=2pt, draw=black!40, fill=white, inner sep=3pt, font=\tiny},
  flow/.style={-{Stealth[length=2mm]}, very thick},
  sflow/.style={-{Stealth[length=1.6mm]}, semithick},
]

\node[font=\small\bfseries, text=black!65] (t1) at (2.0,6.98)  {Multi-modal Inputs};
\node[font=\small\bfseries, text=cEnc!85] (t2) at (8.75,6.98) {Shared Omni Encoder};
\node[font=\small\bfseries, text=black!65] (t3) at (14.2,6.98) {Contrastive Objective};

\node[mbox=cImg] (img) at (2.0,5.9) {};
\draw[cImg, thick, fill=cImg!25] (1.78,6.02) rectangle (2.22,6.42);
\draw[cImg, thin] (2.0,6.02) -- (2.0,6.42) (1.78,6.22) -- (2.22,6.22);
\node at (2.0,5.55) {\textbf{\textcolor{cImg}{Image}}\\[0.5pt]{\tiny patch $14{\times}14$}};

\node[mbox=cVid] (vid) at (2.0,4.4) {};
\foreach \d in {0,0.07,0.14}
  \draw[cVid, thick, fill=cVid!25] ({1.74+\d},{4.56+\d}) rectangle ({2.16+\d},{4.88+\d});
\node at (2.0,4.05) {\textbf{\textcolor{cVid}{Video}}\\[0.5pt]{\tiny tubelet $2{\times}14{\times}14$}};

\node[mbox=cAud] (aud) at (2.0,2.9) {};
\foreach \k/\h in {0/0.05,1/0.12,2/0.20,3/0.26,4/0.15,5/0.22,6/0.09,7/0.18,8/0.25,9/0.13,10/0.21,11/0.08,12/0.04}
  \draw[cAud, line width=1.6pt, line cap=round] ({1.36+\k*0.107},{3.2-\h}) -- ({1.36+\k*0.107},{3.2+\h});
\node at (2.0,2.55) {\textbf{\textcolor{cAud}{Audio}}\\[0.5pt]{\tiny 128-mel $\to$ Conv1D audio tokenizer}};

\node[mbox=cVA] (va) at (2.0,1.4) {};
\foreach \k/\c in {0/cVid,1/cVA,2/cVid,3/cVA,4/cVid,5/cVA,6/cVid,7/cVA}
  \fill[\c!55] ({1.42+\k*0.145},1.6) rectangle ++(0.13,0.26);
\node at (2.0,1.05) {\textbf{\textcolor{cVA}{Video\,+\,Audio}}\\[0.5pt]{\tiny 4 V + 4 A segments, interleaved}};

\node[rounded corners=2pt, draw=cEnc!75, fill=cEnc!4, very thick,
      minimum width=4.0cm, minimum height=6.1cm] (vit) at (7.6,3.65) {};
\fill[cImg] (5.68,6.50) rectangle (6.64,6.62);
\fill[cVid] (6.64,6.50) rectangle (7.60,6.62);
\fill[cAud] (7.60,6.50) rectangle (8.56,6.62);
\fill[cVA]  (8.56,6.50) rectangle (9.52,6.62);
\node[font=\small\bfseries, text=cEnc!95] at (7.6,6.22) {SigLIP-2 ViT (SO400M)};
\node[lbl] at (7.6,5.88) {fully shared across modalities};
\foreach \y in {5.2,4.75,4.3}
  \draw[cEnc!50, fill=cEnc!6] (6.05,\y-0.16) rectangle (9.15,\y+0.16);
\node[font=\tiny, text=cEnc!80] at (7.6,5.2) {Transformer block};
\node[font=\tiny, text=black!60] at (9.4,4.75) {$\times L$};
\node at (7.6,3.85) {\vdots};
\node[rounded corners=1.5pt, draw=cEnc!55, fill=white, thick,
      minimum width=3.3cm, minimum height=0.85cm] at (7.6,2.2)
      {\textbf{3D RoPE}\\[0.5pt]{\tiny decoupled T/H/W freqs., $\theta_t{=}10^4$, $\theta_{h,w}{=}10^3$}};
\node[lbl] at (7.6,1.05) {init.\ SigLIP-2 SO400M image-text weights};

\node[rounded corners=1.5pt, draw=cEnc!70, fill=cEnc!8, very thick,
      minimum width=1.9cm, minimum height=1.0cm] (pool) at (10.95,3.65)
      {\textbf{Varlen}\\\textbf{AttnPool}};
\draw[sflow, cEnc!60] (vit.east) -- (pool.west);

\node[rounded corners=2pt, draw=black!55, fill=black!8, very thick,
      minimum width=3.7cm, minimum height=1.1cm] (txt) at (14.2,5.9)
      {\textbf{SigLIP2 Text Tower}\\[0.5pt]{\tiny caption input (lowercased)}};

\foreach \i/\c in {0/cImg,1/cVid,2/cAud,3/cVA} {
  \foreach \j in {0,1,2,3}
    \fill[\c!16] ({13.36+0.42*\j}, {4.49-0.42*\i}) rectangle ++(0.42,-0.42);
  \fill[\c!80] ({13.36+0.42*\i}, {4.49-0.42*\i}) rectangle ++(0.42,-0.42);
}
\draw[black!45, thin, step=0.42] (13.36,2.81) grid (15.04,4.49);
\node[font=\tiny\bfseries, text=cImg] at (13.16,4.28) {I};
\node[font=\tiny\bfseries, text=cVid] at (13.16,3.86) {V};
\node[font=\tiny\bfseries, text=cAud] at (13.16,3.44) {A};
\node[font=\tiny\bfseries, text=cVA]  at (13.16,3.02) {VA};
\node[lbl] at (14.2,4.62) {Text batch};
\coordinate (matNW) at (12.95,4.85);
\coordinate (matSE) at (15.35,2.6);

\draw[sflow, cImg] (11.93,3.95) to[out=0,in=180] (13.33,4.28);
\draw[sflow, cVid] (11.93,3.78) to[out=0,in=180] (13.33,3.86);
\draw[sflow, cAud] (11.93,3.52) to[out=0,in=180] (13.33,3.44);
\draw[sflow, cVA]  (11.93,3.35) to[out=0,in=180] (13.33,3.02);
\draw[sflow, black!55] (txt.south) -- (14.2,4.56);

\node[rounded corners=2pt, draw=cLoss, fill=cLoss!10, very thick,
      minimum width=3.9cm, minimum height=1.05cm] (loss) at (14.2,1.35)
      {\textbf{\textcolor{cLoss}{Sigmoid Contrastive Loss}}\\[0.5pt]{\tiny SigLIP pairwise logistic loss}};
\draw[sflow, cLoss] (14.2,2.79) -- (loss.north);

\draw[flow, cImg] (img.east) -- (vit.west |- img.east);
\draw[flow, cVid] (vid.east) -- (vit.west |- vid.east);
\draw[flow, cAud] (aud.east) -- (vit.west |- aud.east);
\draw[flow, cVA]  (va.east)  -- (vit.west |- va.east);

\node[rounded corners=2pt, draw=black!35, fill=black!3,
      minimum width=16.6cm, minimum height=0.5cm] (band) at (8.4,0.05) {};
\node[font=\scriptsize\bfseries, text=black!60] at (1.1,0.05) {Training};
\node[pill] at (3.95,0.05)
  {\textcolor{cImg}{\textbf{I}}$\to$\textcolor{cVid}{\textbf{V}}$\to$\textcolor{cAud}{\textbf{A}}$\to$\textcolor{cVA}{\textbf{VA}} round-robin};
\node[pill] at (7.6,0.05)  {4-step grad.\ accumulation $\to$ unified update};
\node[pill] at (11.3,0.05) {256$\times$ MI308X, ZeRO-1, bf16, FA2};
\node[pill] at (14.75,0.05){LAMB lr $2{\times}10^{-5}$, cosine, clip 4.0};

\begin{pgfonlayer}{background}
  \draw[rounded corners=4pt, draw=black!22, fill=black!2] (0.10,0.50) rectangle (4.00,7.25);
  \draw[rounded corners=4pt, draw=cEnc!28, fill=cEnc!2] (5.46,0.50) rectangle (12.04,7.25);
  \draw[rounded corners=4pt, draw=black!22, fill=black!2] (12.15,0.50) rectangle (16.70,7.25);
\end{pgfonlayer}

\end{tikzpicture}%
}
\caption{Overview of Omni-Encoder. One SigLIP-2 ViT backbone is shared across image, video, audio, and audio-visual inputs. The framework is organized along the path from input to training: modality-native input tokenization (so heterogeneous signals enter the shared space without distortion), absolute-timestamp multimodal positional alignment (so visual frames and audio segments live on the same temporal axis), and alternating-modality training with cross-modality gradient accumulation (so every update is balanced across modalities).}
\label{fig:overview}
\end{figure*}



%% file: sec/Figures/fig_radar_tikz.tex
\begin{figure}[!t]
\centering
\begin{tikzpicture}[font=\scriptsize, every node/.style={align=center}]

\definecolor{cOmni}{HTML}{C0392B}
\definecolor{cMiCo}{HTML}{FF6F00}
\definecolor{cQwen}{HTML}{C39BD3}
\definecolor{cLB}{HTML}{2471A3}
\definecolor{cIB}{HTML}{17A589}
\definecolor{cOP}{HTML}{566573}
\definecolor{cSig}{HTML}{D4AC0D}
\definecolor{cVP}{HTML}{E74C3C}
\definecolor{cIV}{HTML}{AF601A}
\definecolor{cM2D}{HTML}{1A5276}
\definecolor{cCLAP}{HTML}{D35400}
\definecolor{cGLAP}{HTML}{196F3D}
\definecolor{cGem}{HTML}{999999}

\def\axmax{{84.10,84.12,78.70,75.47,60.48,76.80,59.62,54.40,63.29,60.60,60.40,34.71,41.70,19.40}}

\def\dOmni{{83.08,84.12,76.97,72.01,54.70,74.26,48.30,54.20,53.50,55.10,60.40,30.70,42.20,19.40}}
\def\dMiCo{{54.12,62.04,48.79,50.69,46.92,56.88,36.20,44.80,38.20,43.90,0,9.40,12.10,8.90}}
\def\dQwen{{72.29,74.92,65.63,64.20,49.08,68.58,59.62,54.40,63.29,51.95,47.35,32.12,0,0}}
\def\dLB{{76.20,77.20,70.10,60.20,37.80,58.40,39.90,44.80,41.00,41.07,37.88,14.95,19.70,16.70}}
\def\dIB{{77.70,59.41,70.93,66.58,49.42,65.92,33.23,36.10,38.70,37.81,29.67,10.60,9.30,6.00}}
\def\dOP{{75.57,68.27,68.12,66.22,48.00,64.70,33.63,39.10,38.70,39.10,29.67,9.97,23.84,12.02}}
\def\dSig{{84.10,79.32,78.70,75.47,55.80,71.70,35.23,41.30,34.51,50.90,35.70,18.80,0,0}}
\def\dVP{{0,0,0,0,0,0,0,51.40,49.60,57.70,0,0,0,0}}
\def\dIV{{0,0,0,0,0,0,31.50,40.70,30.70,49.50,0,0,0,0}}
\def\dMtD{{0,0,0,0,0,0,0,0,0,0,0,0,27.40,10.50}}
\def\dCLAP{{0,0,0,0,0,0,0,0,0,0,0,0,35.40,17.70}}
\def\dGLAP{{0,0,0,0,0,0,0,0,0,0,0,0,41.70,19.40}}
\def\dGem{{81.46,83.46,75.58,70.01,60.48,76.80,49.50,54.00,62.96,60.60,51.48,34.71,6.04,5.74}}

\def\rot{25}

\foreach \i in {0,...,13} {
  \pgfmathsetmacro{\ang}{90-\rot-\i*360/14}
  \pgfmathsetmacro{\rOmni}{\dOmni[\i]/\axmax[\i]*2.2}\coordinate (pOmni\i) at (\ang:\rOmni);
  \pgfmathsetmacro{\rMiCo}{\dMiCo[\i]/\axmax[\i]*2.2}\coordinate (pMiCo\i) at (\ang:\rMiCo);
  \pgfmathsetmacro{\rQwen}{\dQwen[\i]/\axmax[\i]*2.2}\coordinate (pQwen\i) at (\ang:\rQwen);
  \pgfmathsetmacro{\rLB}{\dLB[\i]/\axmax[\i]*2.2}\coordinate (pLB\i) at (\ang:\rLB);
  \pgfmathsetmacro{\rIB}{\dIB[\i]/\axmax[\i]*2.2}\coordinate (pIB\i) at (\ang:\rIB);
  \pgfmathsetmacro{\rOP}{\dOP[\i]/\axmax[\i]*2.2}\coordinate (pOP\i) at (\ang:\rOP);
  \pgfmathsetmacro{\rSig}{\dSig[\i]/\axmax[\i]*2.2}\coordinate (pSig\i) at (\ang:\rSig);
  \pgfmathsetmacro{\rVP}{\dVP[\i]/\axmax[\i]*2.2}\coordinate (pVP\i) at (\ang:\rVP);
  \pgfmathsetmacro{\rIV}{\dIV[\i]/\axmax[\i]*2.2}\coordinate (pIV\i) at (\ang:\rIV);
  \pgfmathsetmacro{\rMtD}{\dMtD[\i]/\axmax[\i]*2.2}\coordinate (pM2D\i) at (\ang:\rMtD);
  \pgfmathsetmacro{\rCLAP}{\dCLAP[\i]/\axmax[\i]*2.2}\coordinate (pCLAP\i) at (\ang:\rCLAP);
  \pgfmathsetmacro{\rGLAP}{\dGLAP[\i]/\axmax[\i]*2.2}\coordinate (pGLAP\i) at (\ang:\rGLAP);
  \pgfmathsetmacro{\rGem}{\dGem[\i]/\axmax[\i]*2.2}\coordinate (pGem\i) at (\ang:\rGem);
}

\foreach \rr in {0.44,0.88,1.32,1.76,2.2}
  \draw[black!25, line width=0.3pt] ({90-\rot}:\rr) \foreach \i in {1,...,13} {-- ({90-\rot-\i*360/14}:\rr)} -- cycle;
\foreach \i in {0,...,13}
  \draw[black!25, line width=0.3pt] (0,0) -- ({90-\rot-\i*360/14}:2.2);

\draw[black!55, densely dashed, line width=0.5pt] (0,0) -- ({0-\rot}:2.42);
\draw[black!55, densely dashed, line width=0.5pt] (0,0) -- ({-51.43-\rot}:2.42);
\draw[black!55, densely dashed, line width=0.5pt] (0,0) -- ({154.29-\rot}:2.42);

\begin{scope}[shift={({25.71-\rot}:3.5)}]
  \draw[black!65, line width=0.6pt] (-0.36,-0.26) rectangle (0.36,0.26);
  \draw[black!65, line width=0.6pt] (-0.36,-0.26) -- (-0.12,0.03) -- (0.03,-0.09) -- (0.15,0.03) -- (0.36,-0.15);
  \fill[cSig] (0.17,0.12) circle (0.055);
\end{scope}
\begin{scope}[shift={({-128.57-\rot}:3.5)}]
  \draw[black!65, line width=0.6pt, rounded corners=2pt] (-0.40,-0.26) rectangle (0.40,0.26);
  \fill[black!65] (-0.11,-0.15) -- (-0.11,0.15) -- (0.15,0.0) -- cycle;
\end{scope}
\begin{scope}[shift={({128.57-\rot}:3.5)}]
  \draw[black!65, line width=1.0pt, line cap=round] (-0.31,-0.09) -- (-0.31,0.09) (-0.20,-0.19) -- (-0.20,0.19) (-0.10,-0.07) -- (-0.10,0.07) (0.00,-0.24) -- (0.00,0.24) (0.10,-0.07) -- (0.10,0.07) (0.20,-0.19) -- (0.20,0.19) (0.31,-0.09) -- (0.31,0.09);
\end{scope}

\fill[cM2D,  fill opacity=0.08] (pM2D0)  \foreach \j in {1,...,13} {-- (pM2D\j)}  -- cycle;
\fill[cCLAP, fill opacity=0.08] (pCLAP0) \foreach \j in {1,...,13} {-- (pCLAP\j)} -- cycle;
\fill[cGLAP, fill opacity=0.08] (pGLAP0) \foreach \j in {1,...,13} {-- (pGLAP\j)} -- cycle;
\fill[cVP,   fill opacity=0.08] (pVP0)   \foreach \j in {1,...,13} {-- (pVP\j)}   -- cycle;
\fill[cIV,   fill opacity=0.08] (pIV0)   \foreach \j in {1,...,13} {-- (pIV\j)}   -- cycle;
\fill[cOmni, fill opacity=0.12] (pOmni0) \foreach \j in {1,...,13} {-- (pOmni\j)} -- cycle;

\draw[cGem,  densely dashed, line width=0.6pt, opacity=0.45] (pGem0)  \foreach \j in {1,...,13} {-- (pGem\j)}  -- cycle;
\draw[cM2D,  line width=0.7pt]                (pM2D0)  \foreach \j in {1,...,13} {-- (pM2D\j)}  -- cycle;
\draw[cCLAP, densely dashed, line width=0.7pt] (pCLAP0) \foreach \j in {1,...,13} {-- (pCLAP\j)} -- cycle;
\draw[cGLAP, line width=0.7pt]                (pGLAP0) \foreach \j in {1,...,13} {-- (pGLAP\j)} -- cycle;
\draw[cIV,   line width=0.8pt]                (pIV0)   \foreach \j in {1,...,13} {-- (pIV\j)}   -- cycle;
\draw[cVP,   densely dashed, line width=0.8pt] (pVP0)   \foreach \j in {1,...,13} {-- (pVP\j)}   -- cycle;
\draw[cSig,  densely dashed, line width=0.8pt] (pSig0)  \foreach \j in {1,...,13} {-- (pSig\j)}  -- cycle;
\draw[cOP,   line width=0.8pt]                (pOP0)   \foreach \j in {1,...,13} {-- (pOP\j)}   -- cycle;
\draw[cIB,   line width=0.8pt]                (pIB0)   \foreach \j in {1,...,13} {-- (pIB\j)}   -- cycle;
\draw[cLB,   line width=0.9pt]                (pLB0)   \foreach \j in {1,...,13} {-- (pLB\j)}   -- cycle;
\draw[cQwen, densely dashed, line width=0.9pt] (pQwen0) \foreach \j in {1,...,13} {-- (pQwen\j)} -- cycle;
\draw[cMiCo, line width=1.0pt]                (pMiCo0) \foreach \j in {1,...,13} {-- (pMiCo\j)} -- cycle;
\draw[cOmni, line width=1.4pt]                (pOmni0) \foreach \j in {1,...,13} {-- (pOmni\j)} -- cycle;

\foreach \i/\t in {
  0/{IN-1K}, 1/{IN-A}, 2/{IN-V2}, 3/{Sketch},
  4/{COCO\\T$\to$I}, 5/{COCO\\I$\to$T},
  6/{DiDeMo}, 7/{MSRVTT}, 8/{ANET}, 9/{VATEX}, 10/{VALOR}, 11/{YouCook2},
  12/{AudioCaps}, 13/{Clotho}}
  \node at ({90-\rot-\i*360/14}:2.8) {\t};

\node[anchor=west, font=\scriptsize\bfseries, align=left] at (-3.7,-3.55) {Omni-Modal};
\node[anchor=west, font=\scriptsize\bfseries, align=left] at (0.35,-3.55) {Specialized \& API};
\foreach \y/\c/\sty/\t in {
  -3.95/cOmni/{}/{Omni-Encoder (Ours)},
  -4.29/cMiCo/{}/{MiCo},
  -4.63/cQwen/{densely dashed}/{Qwen3-VL-2B},
  -4.97/cLB/{}/{LanguageBind},
  -5.31/cIB/{}/{ImageBind},
  -5.65/cOP/{}/{ONE-PEACE}}
{
  \draw[\c, \sty, line width=0.9pt] (-3.7,\y) -- (-3.0,\y);
  \node[anchor=west, align=left] at (-2.85,\y) {\t};
}
\foreach \y/\c/\sty/\t in {
  -3.95/cSig/{densely dashed}/{SigLIP2},
  -4.29/cVP/{densely dashed}/{VideoPrism-B},
  -4.63/cIV/{}/{InternVideo-L},
  -4.97/cM2D/{}/{M2D-CLAP},
  -5.31/cCLAP/{densely dashed}/{CLAP-Fusion},
  -5.65/cGLAP/{}/{GLAP}}
{
  \draw[\c, \sty, line width=0.9pt] (0.35,\y) -- (1.05,\y);
  \node[anchor=west, align=left] at (1.2,\y) {\t};
}
\draw[cGem, densely dashed, opacity=0.6, line width=0.9pt] (0.35,-5.99) -- (1.05,-5.99);
\node[anchor=west, align=left] at (1.2,-5.99) {Gemini Emb. (API)};

\end{tikzpicture}
\caption{Per-axis normalized R@1 across 14 benchmarks spanning image classification, image-text retrieval, video-text retrieval, and audio-text retrieval. With a \emph{single} set of shared parameters, our Omni-Encoder (red, solid) maintains balanced performance on every axis, whereas modality-specific encoders (dashed) excel only in their target domain.}
\label{fig:radar}
\end{figure}

%% file: sec/2_related_work.tex
\section{Related Work}
\label{sec:related_work}

Our work relates to four lines of research: image-text contrastive learning, video-language modeling with temporal position encoding, audio-language representation learning, and unified multimodal encoders.

\subsection{Image-Text Contrastive Learning}
\label{sec:rw_image_text}

Dual-encoder contrastive pretraining on large-scale noisy image-text pairs has become the dominant paradigm for transferable visual representations~\cite{radford2021learning,jia2021align}.
Subsequent work improves this paradigm along several axes:
(i)~loss design, replacing the softmax objective with pairwise sigmoid losses for more scalable training~\cite{zhai2023sigmoid,tschannen2025siglip2};
(ii)~training recipes and model scaling~\cite{sun2023evaclip,zhai2022lit,pham2023basic};
(iii)~data curation and reproducible scaling studies~\cite{gadre2024datacomp,cherti2023reproducible};
and (iv)~supervision beyond pure contrast, such as captioning or bootstrapped objectives~\cite{li2022blip,yu2022coca}.
Nearly all of these encoders adopt a patch-based Vision Transformer backbone~\cite{dosovitskiy2021image}, a formulation shared with self-supervised visual encoders~\cite{oquab2024dinov2} and naturally extensible beyond static images.
Omni-Encoder inherits the sigmoid contrastive recipe and the ViT backbone from this line, but differs in that a single parameter set is extended from image-text to video, audio, and audio-visual inputs, rather than remaining an image-text specialist.


\subsection{Video-Language Modeling and Temporal Position Encoding}
\label{sec:rw_video}

Extending image-text pretraining to video requires modeling temporal dynamics, and existing work follows two routes.
(i)~One route adapts pretrained image encoders to video through frame sampling with temporal pooling or lightweight temporal modules~\cite{luo2022clip4clip,bain2021frozen,xu2021videoclip,ni2022expanding,rasheed2023fine,wang2021actionclip,lin2022egovlp}.
(ii)~The other builds video-native foundation encoders with dedicated video pretraining objectives~\cite{wang2022internvideo,zhao2024videoprism,li2023umt}.
In both routes, temporal position is typically encoded with frame-index-based schemes, ranging from 2D/3D rotary embeddings adopted in recent multimodal systems~\cite{su2024roformer,arnab2021vivit,wang2024qwen2,zhang2024llavavideo,bai2025qwen2} to length-extrapolation and interpolation techniques developed for long-context modeling~\cite{press2022train,chen2023extending,peng2024yarn}.
Frame indices, however, conflate temporal position with sampling rate and carry no absolute physical time, which limits their use for audio-visual synchronization.
This limitation is amplified under resampling: the same content sampled at different frame rates receives entirely different position indices, and the learned temporal positions do not transfer across rates.
Remedies based on index interpolation can stretch the index grid to longer sequences, but the grid itself remains rate-dependent rather than grounded in physical time.
Absolute timestamps remove this coupling: a video frame at $2$\,s and an audio segment centered at $2$\,s share the same temporal coordinate, regardless of how densely each modality is sampled.
Omni-Encoder differs in that the temporal axis of its multimodal rotary position embedding is parameterized by absolute timestamps, so visual frames and audio segments share one spatiotemporal coordinate system within the same backbone.


\subsection{Audio-Language Representation Learning}
\label{sec:rw_audio}

Audio understanding has evolved from supervised classification on large annotated corpora~\cite{gemmeke2017audioset,kong2020panns} to self-supervised pretraining over waveforms or spectrograms~\cite{baevski2020wav2vec,hsu2021hubert,gong2021ast,gong2022ssast,huang2022audiomae,chen2023beats,radford2023whisper}, and more recently to contrastive language-audio pretraining~\cite{elizalde2023clap,wu2023clap} supported by weakly labeled or synthetic caption corpora~\cite{mei2024wavcaps,kim2019audiocaps}.
Two observations recur across this literature:
(i)~front-ends that respect audio signal statistics---temporal convolutions or spectrogram-specific patching---consistently outperform directly re-using image patchifiers~\cite{radford2023whisper,gong2021ast,guzhov2022audioclip};
(ii)~once inputs are properly tokenized, the downstream semantic processing is largely compatible with standard Transformer backbones.
Omni-Encoder follows these observations: a modality-native 1D-convolutional tokenizer converts mel-spectrograms into tokens aligned with visual patches, so audio can be processed by the same shared backbone without a private encoder.


\subsection{Unified Multimodal Encoders}
\label{sec:rw_unified}

Unifying multiple modalities in one model has been explored along three directions.
(i)~Binding-based approaches align independently trained modality encoders into a common space through an anchor modality~\cite{girdhar2023imagebind,zhu2024languagebind}, which enables cross-modal retrieval but keeps per-modality parameters disjoint and the alignment post-hoc.
(ii)~Architecture-sharing approaches route modalities through partially shared backbones while retaining modality-specific experts or adapters~\cite{wang2023beit3,bao2024onepeace,akbari2021vatt,jaegle2021perceiver,zhang2023metatransformer,han2024onellm}, or embed frozen encoders into larger multimodal systems~\cite{lu2022unified,comanici2025gemini}.
(iii)~Audio-visual joint learning exploits natural audio-visual co-occurrence as a self-supervision signal~\cite{arandjelovic2017look,morgado2021avid,gong2023cavmae,shi2022learning}, but remains largely confined to the audio-visual pair rather than general image, video, and audio understanding.
Omni-Encoder differs from all three directions in that a single fully shared backbone serves image, video, audio, and audio-visual inputs with end-to-end joint optimization, and audio-visual temporal correspondence is encoded inside the position encoding rather than added by an external alignment stage.
Table~\ref{tab:related_comparison} summarizes the comparison.


\begin{table}[t]
\centering
\caption{Comparison with representative unified multimodal encoders.}
\label{tab:related_comparison}
\resizebox{\linewidth}{!}{
\begin{tabular}{lccccc}
\toprule
\textbf{Method} & \textbf{Modalities} & \textbf{Shared} & \textbf{End-to-} & \textbf{AV} & \textbf{Per-mod.} \\
 & & \textbf{backbone} & \textbf{end} & \textbf{align.} & \textbf{SOTA} \\
\midrule
CLIP~\cite{radford2021learning} & I, T & $\checkmark$ & $\checkmark$ & $\times$ & $\checkmark$ \\
ImageBind~\cite{girdhar2023imagebind} & I, V, A, + & $\times$ & $\times$ & $\times$ & $\times$ \\
LanguageBind~\cite{zhu2024languagebind} & I, V, A, + & $\times$ & $\times$ & $\times$ & $\times$ \\
ONE-PEACE~\cite{bao2024onepeace} & I, A, T & partial & $\checkmark$ & $\times$ & $\times$ \\
BEiT-3~\cite{wang2023beit3} & I, T & partial & $\checkmark$ & $\times$ & $\checkmark$ \\
\midrule
\textbf{Omni-Encoder} & I, V, A, VA & $\checkmark$ & $\checkmark$ & $\checkmark$ & $\checkmark$ \\
\bottomrule
\end{tabular}
}
\vspace{-10pt}
\end{table}


%% file: sec/3_method.tex
\section{Methodology}
\label{sec:method}

Omni-Encoder processes image, video, audio, and audio-visual inputs within a single SigLIP-2 ViT backbone~\cite{tschannen2025siglip2}, organized along the path from input to training (Figure~\ref{fig:overview}).
Modality-native tokenizers first map each raw signal into a shared embedding space $\mathbb{R}^d$: images and videos through a 3D convolutional front-end~\cite{arnab2021vivit}, and audio mel-spectrograms through a dedicated 1D convolutional tokenizer (Sec.~\ref{sec:audio_tokenizer}).
All tokens are then equipped with a multimodal rotary position embedding parameterized by absolute timestamps (Sec.~\ref{sec:tmrope}), which places visual frames and audio segments in a unified spatiotemporal coordinate system; for audio-visual inputs, video and audio tokens are further interleaved at the segment level so that cross-modal attention occurs within a single forward pass (Sec.~\ref{sec:va_training}).
The model is trained end-to-end with a sigmoid contrastive objective (Sec.~\ref{sec:objective}) under an alternating-modality schedule with cross-modality gradient accumulation (Sec.~\ref{sec:training_strategy}); the default configuration is summarized in Table~\ref{tab:impl}.


\subsection{Shared Backbone, Native Tokenizers}
\label{sec:arch}

A single SigLIP-2 ViT backbone~\cite{tschannen2025siglip2,dosovitskiy2021image}, initialized from pretrained image-text weights, processes all four modalities.
The motivation is that objects, scenes, events, and sounds share high-level semantic structure: recognizing ``a dog barking'' in audio and ``a dog running'' in video both require the concept of ``dog.''
A shared Transformer~\cite{vaswani2017attention} captures these modality-agnostic semantics and enforces cross-modal alignment from the first layer---an \emph{early fusion} at the encoder level, in contrast to the \emph{late fusion} of multimodal large language models~\cite{liu2023visual,wang2024qwen2}, where modality-specific encoders extract features independently before fusion in a language decoder.
A single set of Transformer weights serving all modalities is also parameter-efficient, reducing storage and simplifying deployment.

While the backbone is shared, the tokenizers must remain modality-native.
Converting all inputs into a common format---e.g., treating mel-spectrograms as images and re-using the visual tokenizer---discards audio-specific statistics (harmonic frequency structure, temporal sparsity, log-compressed amplitude) before any semantic processing occurs.
Omni-Encoder therefore tokenizes images and videos with a shared 3D-CNN front-end~\cite{arnab2021vivit}, where consecutive frame pairs are processed jointly for videos and a single frame is duplicated for images, while audio uses the dedicated 1D tokenizer described in Sec.~\ref{sec:audio_tokenizer}.
After projection into $\mathbb{R}^d$, all tokens are processed identically by the shared Transformer, achieving unified semantic processing without sacrificing modality-specific signal fidelity.


\subsection{Multimodal Rotary Position Embedding}
\label{sec:tmrope}

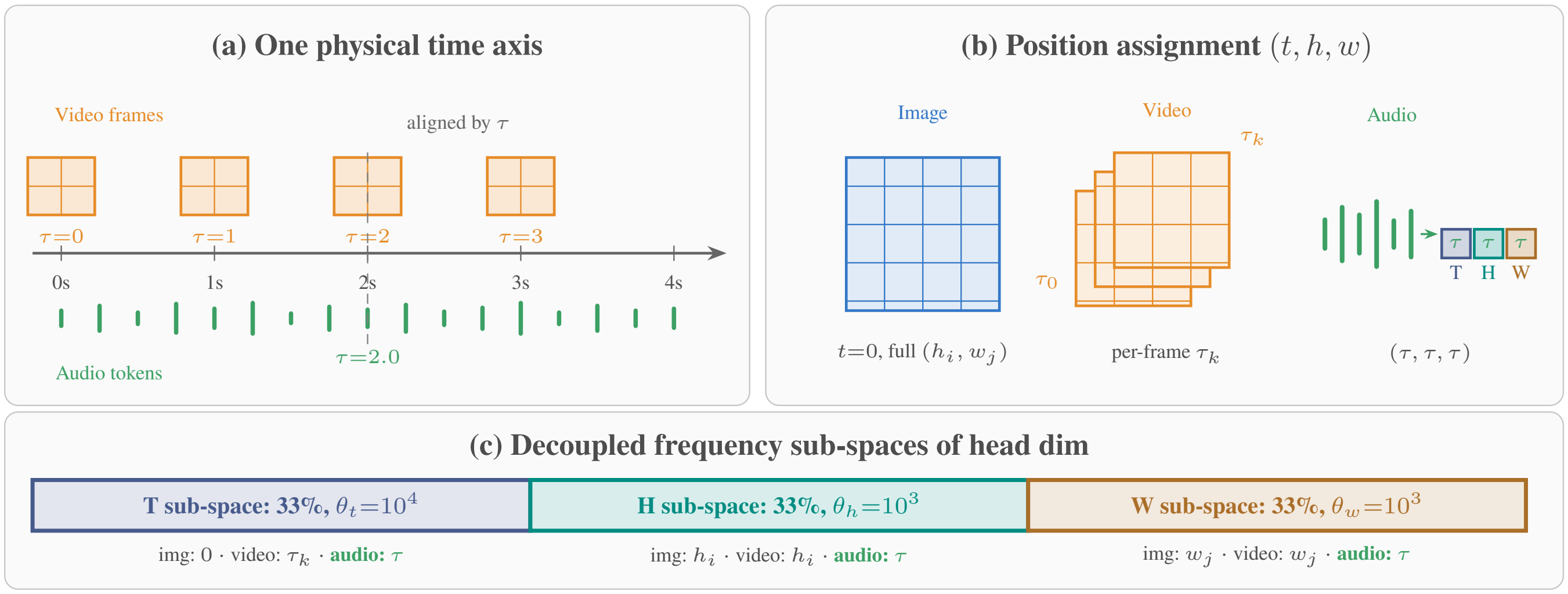
\begin{figure*}[t]
\centering
\definecolor{cImg}{RGB}{52,120,200}
\definecolor{cVid}{RGB}{230,140,40}
\definecolor{cAud}{RGB}{60,160,100}
\definecolor{cT}{RGB}{70,90,140}
\definecolor{cH}{RGB}{0,140,130}
\definecolor{cW}{RGB}{170,110,40}
\resizebox{\textwidth}{!}{%
\begin{tikzpicture}[
  font=\scriptsize,
  every node/.style={align=center, inner sep=1.5pt},
  lbl/.style={font=\tiny, text=black!70},
]

\node[font=\small\bfseries, text=black!70] at (4.3,6.25) {(a) One physical time axis};

\draw[-{Stealth[length=2mm]}, thick, black!60] (0.7,4.1) -- (7.95,4.1);
\foreach \t in {0,1,2,3,4} {
  \draw[black!60] ({1.0+1.6*\t},4.02) -- ({1.0+1.6*\t},4.18);
  \node[lbl] at ({1.0+1.6*\t},3.8) {\t s};
}

\node[lbl, text=cVid] at (1.5,5.55) {Video frames};
\foreach \t in {0,1,2,3} {
  \draw[cVid, thick, fill=cVid!20] ({1.0+1.6*\t-0.35},4.5) rectangle ({1.0+1.6*\t+0.35},5.1);
  \draw[cVid, thin] ({1.0+1.6*\t},4.5) -- ({1.0+1.6*\t},5.1)
                     ({1.0+1.6*\t-0.35},4.8) -- ({1.0+1.6*\t+0.35},4.8);
  \node[lbl, text=cVid] at ({1.0+1.6*\t},4.28) {$\tau{=}\t$};
}

\node[lbl, text=cAud] at (1.5,2.85) {Audio tokens};
\foreach \k/\h in {0/0.08,1/0.13,2/0.06,3/0.15,4/0.10,5/0.16,6/0.05,7/0.12,8/0.09,9/0.14,10/0.07,11/0.11,12/0.16,13/0.06,14/0.13,15/0.08,16/0.10}
  \draw[cAud, line width=1.3pt, line cap=round] ({1.0+\k*0.4},{3.42-\h}) -- ({1.0+\k*0.4},{3.42+\h});

\draw[dashed, black!50] (4.2,3.15) -- (4.2,5.25);
\node[lbl, text=cAud] at (4.2,3.02) {$\tau{=}2.0$};
\node[lbl, text=black!60] at (5.15,5.45) {aligned by $\tau$};

\node[font=\small\bfseries, text=black!70] at (12.55,6.25) {(b) Position assignment $(t,h,w)$};

\node[lbl, text=cImg] at (10.0,5.55) {Image};
\fill[cImg!15] (9.2,3.5) rectangle (10.8,5.1);
\draw[cImg, thin, step=0.4] (9.2,3.5) grid (10.8,5.1);
\draw[cImg, thick] (9.2,3.5) rectangle (10.8,5.1);
\node[lbl] at (10.0,3.05) {$t{=}0$, full $(h_i,w_j)$};

\node[lbl, text=cVid] at (12.55,5.6) {Video};
\foreach \d in {0,0.2,0.4} {
  \fill[cVid!15] ({11.6+\d},{3.55+\d}) rectangle ({12.8+\d},{4.75+\d});
  \draw[cVid, thin, step=0.4] ({11.6+\d},{3.55+\d}) grid ({12.8+\d},{4.75+\d});
  \draw[cVid, thick] ({11.6+\d},{3.55+\d}) rectangle ({12.8+\d},{4.75+\d});
}
\node[lbl, text=cVid] at (11.3,3.8) {$\tau_0$};
\node[lbl, text=cVid] at (13.45,5.3) {$\tau_k$};
\node[lbl] at (12.55,3.05) {per-frame $\tau_k$};

\node[lbl, text=cAud] at (14.9,5.55) {Audio};
\foreach \k/\h in {0/0.15,1/0.28,2/0.20,3/0.34,4/0.14,5/0.24}
  \draw[cAud, line width=1.4pt, line cap=round] ({14.2+\k*0.18},{4.3-\h}) -- ({14.2+\k*0.18},{4.3+\h});
\draw[-{Stealth[length=1.4mm]}, cAud, semithick] (15.2,4.3) -- (15.38,4.3);
\foreach \s/\c in {0/cT,1/cH,2/cW}
  \draw[\c, thick, fill=\c!25] ({15.42+\s*0.34},4.05) rectangle ++(0.3,0.3);
\node[font=\tiny, text=cAud] at (15.57,4.2) {$\tau$};
\node[font=\tiny, text=cAud] at (15.91,4.2) {$\tau$};
\node[font=\tiny, text=cAud] at (16.25,4.2) {$\tau$};
\node[font=\tiny, text=cT] at (15.57,3.9) {T};
\node[font=\tiny, text=cH] at (15.91,3.9) {H};
\node[font=\tiny, text=cW] at (16.25,3.9) {W};
\node[lbl] at (15.3,3.05) {$(\tau,\tau,\tau)$};

\node[font=\small\bfseries, text=black!70] at (8.5,2.07) {(c) Decoupled frequency sub-spaces of head dim};

\draw[cT, very thick, fill=cT!15] (0.7,1.2)  rectangle (5.9,1.73);
\draw[cH, very thick, fill=cH!15] (5.9,1.2)  rectangle (11.1,1.73);
\draw[cW, very thick, fill=cW!15] (11.1,1.2) rectangle (16.3,1.73);
\node[font=\scriptsize\bfseries, text=cT] at (3.3,1.465)  {T sub-space: 33\%, $\theta_t{=}10^4$};
\node[font=\scriptsize\bfseries, text=cH] at (8.5,1.465) {H sub-space: 33\%, $\theta_h{=}10^3$};
\node[font=\scriptsize\bfseries, text=cW] at (13.7,1.465){W sub-space: 33\%, $\theta_w{=}10^3$};

\node[lbl] at (3.3,0.93)   {img: $0$ $\cdot$ video: $\tau_k$ $\cdot$ \textcolor{cAud}{\textbf{audio: $\tau$}}};
\node[lbl] at (8.5,0.93)  {img: $h_i$ $\cdot$ video: $h_i$ $\cdot$ \textcolor{cAud}{\textbf{audio: $\tau$}}};
\node[lbl] at (13.7,0.93) {img: $w_j$ $\cdot$ video: $w_j$ $\cdot$ \textcolor{cAud}{\textbf{audio: $\tau$}}};

\coordinate (aNW) at (0.55,6.55);  \coordinate (aSE) at (8.05,2.65);
\coordinate (bNW) at (8.5,6.55);   \coordinate (bSE) at (16.55,2.65);
\coordinate (cNW) at (0.55,2.3);   \coordinate (cSE) at (16.55,0.73);
\begin{pgfonlayer}{background}
  \node[rounded corners=4pt, draw=black!22, fill=black!2, inner sep=4pt, fit=(aNW)(aSE)] {};
  \node[rounded corners=4pt, draw=black!22, fill=black!2, inner sep=4pt, fit=(bNW)(bSE)] {};
  \node[rounded corners=4pt, draw=black!22, fill=black!2, inner sep=4pt, fit=(cNW)(cSE)] {};
\end{pgfonlayer}

\end{tikzpicture}%
}
\caption{Multimodal 3D RoPE with absolute timestamps. (a)~Video frames and audio tokens share one physical time axis: each token's temporal coordinate is its absolute timestamp in seconds, so different sampling rates become commensurable and co-occurring content is aligned. (b)~Modality-specific position assignment: images use $t{=}0$ over the full $(h,w)$ grid; video frames use per-frame timestamps $(\tau_k, h_i, w_j)$; audio replicates its timestamp into all three sub-spaces $(\tau,\tau,\tau)$. (c)~The attention head dimension is split into decoupled T/H/W frequency sub-spaces ($33/33/33\%$, $\theta_t{=}10^4$, $\theta_h{=}\theta_w{=}10^3$); audio fills all three with $\tau$ so the full rotary bandwidth encodes time.}
\label{fig:rope}
\end{figure*}


Standard SigLIP-2 employs 2D RoPE~\cite{su2024roformer}, which encodes only spatial positions $(H, W)$: all video frames would receive identical encodings, erasing temporal structure, and audio, a purely temporal signal, has no meaningful place in a spatial grid.
We therefore extend RoPE to 3D $(T, H, W)$ with a key design choice (Figure~\ref{fig:rope}): the temporal axis is parameterized by \emph{absolute timestamps} (in seconds) rather than integer frame indices.
Timestamps align modalities with fundamentally different sampling rates (e.g., 16\,kHz audio vs.\ 1\,fps video) onto one physical time axis, and decouple temporal position from sampling density, so the model generalizes across frame rates and durations without retraining.
The assignment is modality-specific: images use $T{=}0$ over the full $(H, W)$ grid; videos use the absolute timestamp of each frame; audio replicates its timestamp across all three subspaces, encoding it as a purely temporal signal.

Continuous timestamp parameterization also brings strong temporal extrapolation.
Although most training videos are 5--30 seconds long, the model generalizes at inference to videos several times longer and to arbitrary frame counts, because it learns temporal relations in physical time intervals rather than discrete frame positions---an advantage analogous to continuous rotary-embedding extrapolation in long-context modeling~\cite{peng2024yarn}.


\subsection{Convolutional Audio Tokenizer}
\label{sec:audio_tokenizer}

Mel-spectrograms differ fundamentally from natural images: the frequency axis carries harmonic structures with non-local correlations (fundamentals and overtones span distant bins), the time axis exhibits event sparsity rather than dense spatial content, and amplitudes are log-compressed.
Audio representation learning has repeatedly shown that respecting these statistics in the front-end improves over directly transferring image architectures to spectrograms~\cite{guzhov2022audioclip,gong2021ast,chen2023beats}.

Inspired by the convolutional front-end of Whisper~\cite{radford2023whisper}, the tokenizer processes mel-spectrograms with 1D convolutions along the temporal axis.
Unlike 2D patch embeddings that fragment frequency bands into independent patches, the 1D design preserves the full frequency spectrum at each time step, letting the filters capture harmonic patterns holistically.
Output tokens are projected into the same $\mathbb{R}^d$ space as visual tokens, so the shared backbone processes audio seamlessly while the dedicated tokenizer preserves audio-specific signal characteristics; token counts and strides are listed in Table~\ref{tab:impl}.


\begin{table}[t]
\centering
\caption{Implementation details of Omni-Encoder.}
\label{tab:impl}
\resizebox{\linewidth}{!}{
\begin{tabular}{ll}
\toprule
\textbf{Component} & \textbf{Configuration} \\
\midrule
Shared backbone & SigLIP-2 ViT-SO400M~\cite{tschannen2025siglip2}, patch $14{\times}14$ \\
Visual tokenizer & 3D-CNN, kernel $2{\times}14{\times}14$ (two-frame tube) \\
Audio tokenizer & 128 mel bins, 1D conv (stride 2), $\le$1024 steps ($\le$512 tokens) \\
Position encoding & $\rho_T{:}\rho_H{:}\rho_W = 0.33{:}0.33{:}0.33$; $\theta_T{=}10^4$, $\theta_H{=}\theta_W{=}10^3$ \\
VA interleaving & 8 frames $\to$ 4 video segments; 4 audio segments; 224 audio steps/segment \\
Optimizer & LAMB~\cite{you2020lamb}, lr $2{\times}10^{-5}$, cosine decay \\
Batch ratio (I:V:A:VA) & 64:16:64:16, accumulated over 4 micro-steps (\texttt{update\_freq}{=}4) \\
Gradient clipping & 4.0 \\
\bottomrule
\end{tabular}
}
\end{table}

\subsection{Audio-Visual Segment Interleaving}
\label{sec:va_training}

For audio-visual inputs, we interleave video and audio tokens at the segment level to enable fine-grained cross-modal alignment inside the shared encoder.
Video frames are grouped by the 3D-CNN tokenizer into temporal segments, and the audio stream is divided into the same number of equal-duration segments; tokens are then ordered as $[\mathbf{V}_1, \mathbf{A}_1, \dots, \mathbf{V}_K, \mathbf{A}_K]$ in a single sequence.
Because the multimodal position embedding assigns absolute timestamps to both modalities within each segment, self-attention can attend across video and audio tokens inside the same temporal window, letting cross-modal correspondences emerge naturally at the representation level~\cite{gong2023cavmae,morgado2021avid}.

A naive alternative concatenates all video tokens before all audio tokens.
This disrupts temporal locality: attention between temporally corresponding $\mathbf{V}_k$ and $\mathbf{A}_k$ must span the whole sequence, and the model cannot easily learn that a sound at time $t$ corresponds to the visual event at the same time.
Interleaving keeps co-occurring events adjacent in the sequence, making correspondence learning significantly easier.


\subsection{Training Objective}
\label{sec:objective}

Omni-Encoder is trained with the sigmoid contrastive objective of SigLIP~\cite{zhai2023sigmoid,tschannen2025siglip2}.
For a micro-batch $\mathcal{B}$ of $N$ modality-text pairs, let $\bar{z}_i$ and $\bar{u}_j$ denote the $\ell_2$-normalized embeddings of the $i$-th modality input and the $j$-th text, produced by the shared backbone followed by a learnable-probe attention pooling head that aggregates variable-length token sequences without padding, and by the text tower, respectively.
The objective applies a per-pair logistic loss over all $N^2$ combinations:
\begin{equation}
\mathcal{L}_{\mathrm{sig}}(\mathcal{B}) = -\frac{1}{N} \sum_{i=1}^{N} \sum_{j=1}^{N} \log \sigma\!\left( y_{ij} \left( s \, \bar{z}_i^{\!\top} \bar{u}_j + b \right) \right),
\label{eq:siglip}
\end{equation}
where $y_{ij}{=}+1$ for matched pairs ($i{=}j$) and $-1$ otherwise, $\sigma(\cdot)$ is the logistic function, and $s$ and $b$ are learnable scale and bias parameters.
Unlike softmax-based InfoNCE, the sigmoid loss decouples each pair from a batch-level softmax denominator, which keeps the objective well-defined when the four modality micro-batches are processed in separate forward-backward passes.

Each training iteration evaluates the objective independently on the four modality micro-batches and accumulates the gradients into a single update:
\begin{equation}
\mathcal{L} \;=\; \sum_{m \in \{\mathrm{I},\,\mathrm{V},\,\mathrm{A},\,\mathrm{VA}\}} \mathcal{L}_{\mathrm{sig}}(\mathcal{B}_m),
\label{eq:total}
\end{equation}
so that one optimizer step always reflects the joint gradient of all modalities.


\subsection{Cross-Modality Gradient Accumulation}
\label{sec:training_strategy}

Training a shared encoder on multiple modalities introduces \emph{modality competition}~\cite{huang2022modality}: when gradients from different modalities are applied independently, modalities with larger gradient magnitudes or faster convergence can dominate the optimization trajectory and cause slower modalities to stagnate~\cite{yu2020pcgrad}.
We address this with an alternating-modality schedule: each optimizer step consists of four sequential forward-backward passes---Image~$\to$~Video~$\to$~Audio~$\to$~VA---whose gradients, computed from Eq.~\eqref{eq:total}, are accumulated before a single unified update.
The optimizer therefore always receives a \emph{balanced} gradient that reflects the joint objective, which reduces gradient variance, avoids the oscillation caused by temporarily dominant modalities, and encourages convergence toward a solution that benefits all modalities.
Compared with gradient-surgery methods such as PCGrad~\cite{yu2020pcgrad}, accumulation is simpler, adds no per-step computation, and scales naturally with the number of modalities.

%% file: sec/4_data_recipe.tex
\section{Data Recipe}
\label{sec:data}

\begin{table}[t]
\centering
\caption{Overview of the four-modality training corpus.}
\label{tab:training_data_overview}
\scriptsize
\setlength{\tabcolsep}{4pt}
\begin{tabular}{@{}l c p{4.2cm}@{}}
\toprule
\textbf{Data} & \textbf{Scale} & \textbf{Key sources} \\
\midrule
Image-text & 170M & DataComp; knowledge-enhanced corpus \\
Video-text & 42M & Mix42M (multi-source) \\
Audio-text & 42M & WavCaps + supplementary sources \\
Audio-visual & 30M & YouTube, AudioSet, VGGSound, YouCook2 \\
\midrule
\multicolumn{3}{@{}l@{}}{Mixing ratio 64:16:64:16 (image:video:audio:audio-visual)} \\
\bottomrule
\end{tabular}
\end{table}

\definecolor{cRankTwo}{RGB}{35,140,70}
\definecolor{cRankThree}{RGB}{198,55,55}
\newcommand{\secondbest}[1]{\textcolor{cRankTwo}{#1}}
\newcommand{\thirdbest}[1]{\textcolor{cRankThree}{#1}}
\begin{table*}[!t]
\centering
\caption{Comprehensive comparison of vision-language models across modality capabilities. We categorize methods into four tiers based on their supported modalities: Image-Text only, Video-Text only, Image+Video-Text, and Omni-modal (Image+Video+Audio+Text). Our Omni-Encoder achieves strong performance across all modalities and tasks. All retrieval metrics are R@1 (\%). \textbf{Bold}, \secondbest{green}, and \thirdbest{red} indicate the best, second-best, and third-best result in each column, respectively.}
\label{tab:main_results}
\resizebox{\textwidth}{!}{%
\begin{tabular}{l|cccc|cc|cc|cccccc|cc}
\toprule
\multirow{3}{*}{\textbf{Method}}
  & \multicolumn{4}{c|}{\textbf{Image Classification}}
  & \multicolumn{2}{c|}{\textbf{Image-Text}}
  & \multicolumn{2}{c|}{\textbf{Audio-Text}}
  & \multicolumn{6}{c|}{\textbf{Video-Text}}
  & \multicolumn{2}{c}{\textbf{Video-Audio-Text}} \\
\cmidrule(lr){2-5} \cmidrule(lr){6-7} \cmidrule(lr){8-9} \cmidrule(lr){10-15} \cmidrule(lr){16-17}
  & \multicolumn{4}{c|}{(Accuracy \%)}
  & \multicolumn{2}{c|}{(R@1 \%)}
  & \multicolumn{2}{c|}{(R@1 \%)}
  & \multicolumn{6}{c|}{(R@1 \%)}
  & \multicolumn{2}{c}{(R@1 \%)} \\
\cmidrule(lr){2-5} \cmidrule(lr){6-7} \cmidrule(lr){8-9} \cmidrule(lr){10-15} \cmidrule(lr){16-17}
  & IN-1K & IN-A & IN-V2 & Sketch
  & COCO T$\to$I & COCO I$\to$T
  & AudioCaps & Clothv2
  & DiDemo & MSRVTT & ANET & YouCook2 & VALOR & VATEX
  & VALOR T$\to$AV & VATEX T$\to$AV \\
\midrule
\multicolumn{17}{l}{\textit{\textbf{Tier 1: Image-Text Models}}} \\
CLIP-ViT-L/14 & 76.20 & 77.20 & 70.10 & 60.20 & 37.80 & 58.40 & - & - & 29.80 & 35.00 & 25.20 & 8.60 & 13.20 & 45.20 & - & - \\
SigLIP-SO400M & \textbf{84.10} & 79.32 & \textbf{78.70} & \textbf{75.47} & \secondbest{55.80} & \thirdbest{71.70} & - & - & 35.23 & 41.30 & 34.51 & 18.80 & 35.70 & 50.90 & - & - \\
Seed-ViT & \secondbest{83.60} & \textbf{85.50} & \secondbest{77.60} & \secondbest{74.10} & - & - & - & - & - & - & - & - & - & - & - & - \\
\midrule
\multicolumn{17}{l}{\textit{\textbf{Tier 2: Video-Text Models}}} \\
Singularity-B & - & - & - & - & - & - & - & - & - & 34.00 & 30.60 & - & - & - & - & - \\
VideoCoCa-g & - & - & - & - & - & - & - & - & - & 43.90 & 34.50 & - & - & \thirdbest{53.20} & - & - \\
InternVideo-L & - & - & - & - & - & - & - & - & 31.50 & 40.70 & 30.70 & - & - & 49.50 & - & - \\
UMT-L & - & - & - & - & - & - & - & - & - & 42.60 & 42.80 & - & - & - & - & - \\
VideoPrism-B & - & - & - & - & - & - & - & - & - & 51.40 & 49.60 & - & - & \textbf{57.70} & - & - \\
\midrule
\multicolumn{17}{l}{\textit{\textbf{Tier 3: Audio-Text Models}}} \\
M2D-CLAP & - & - & - & - & - & - & 27.40 & 10.50 & - & - & - & - & - & - & - & - \\
MS-CLAP & - & - & - & - & - & - & 23.40 & \secondbest{17.80} & - & - & - & - & - & - & - & - \\
CLAP-Fusion & - & - & - & - & - & - & \thirdbest{35.40} & \thirdbest{17.70} & - & - & - & - & - & - & - & - \\
GLAP & - & - & - & - & - & - & \secondbest{41.70} & \textbf{19.40} & - & - & - & - & - & - & - & - \\
\midrule
\multicolumn{17}{l}{\textit{\textbf{Tier 4: Omni-Modal Models (Image+Video+Audio+Text)}}} \\
ImageBind & 77.70 & 59.41 & 70.93 & 66.58 & 49.42 & 65.92 & 9.30 & 6.00 & 33.23 & 36.10 & 38.70 & 10.60 & 29.67 & 37.81 & - & - \\
LanguageBind & 76.20 & 77.20 & 70.10 & 60.20 & 37.80 & 58.40 & 19.70 & 16.70 & 39.90 & 44.80 & 41.00 & 14.95 & 37.88 & 41.07 & - & - \\
MiCo & 54.12 & 62.04 & 48.79 & 50.69 & 46.92 & 56.88 & 12.10 & 8.90 & 36.20 & 44.80 & 38.20 & 9.40 & \textbf{67.23} & 43.90 & \textbf{73.38} & \thirdbest{37.58} \\
ONE-PEACE & 75.57 & 68.27 & 68.12 & 66.22 & 48.00 & 64.70 & 23.84 & 12.02 & 33.63 & 39.10 & 38.70 & 9.97 & 29.67 & 39.10 & 26.62 & 27.06 \\
Qwen3-VL-Embedding-2B & 72.29 & 74.92 & 65.63 & 64.20 & 49.08 & 68.58 & - & - & \textbf{59.62} & \textbf{54.40} & \textbf{63.29} & \secondbest{32.12} & 47.35 & 51.95 & - & - \\
Gemini-Omni-Embedding & 81.46 & \thirdbest{83.46} & 75.58 & 70.01 & \textbf{60.48} & \textbf{76.80} & 5.65 & 5.74 & \secondbest{49.50} & \thirdbest{54.00} & \secondbest{62.96} & \textbf{34.71} & \thirdbest{48.90} & - & \thirdbest{51.48} & \secondbest{60.60} \\
\textbf{Omni-Encoder (Ours)} & \thirdbest{83.08} & \secondbest{84.12} & \thirdbest{76.97} & \thirdbest{72.01} & \thirdbest{54.70} & \secondbest{74.26} & \textbf{42.20} & \textbf{19.40} & \thirdbest{48.30} & \secondbest{54.20} & \thirdbest{53.50} & \thirdbest{30.70} & \secondbest{60.40} & \secondbest{55.10} & \secondbest{68.50} & \textbf{72.40} \\
\bottomrule
\end{tabular}%
}
\end{table*}

Our training corpus spans four modalities---image-text, video-text, audio-text, and temporally aligned audio-visual data (Table~\ref{tab:training_data_overview})---mixed at a 64:16:64:16 sample ratio per optimizer update.
Training alternates single-modality batches in this order and accumulates their gradients into one joint update, so each update sees all four modalities at an exact, deterministic ratio rather than a stochastic sampling probability.
Pure batches further avoid padding across heterogeneous input shapes and keep per-micro-batch token budgets---and thus peak memory---nearly uniform.
Image- and audio-text pairs receive the highest weights, as they are the most abundant and cleanly annotated, while video and audio-visual data are down-weighted due to noisier annotations and higher per-sample cost---though their multi-frame, multi-segment inputs inflate per-sample token counts enough to largely offset the lower sample weights, yielding near-balanced token shares per update; all four share one curation pipeline of quality filtering, category-balanced sampling, and MLLM-based recaptioning.

\noindent\textbf{Image-text.}\label{sec:data_image}
We sample 150M semantically balanced pairs from DataComp~\cite{gadre2024datacomp} ($\sim$710M) via cluster-based sampling, where $k$-means clusters over image embeddings are rebalanced so that head and tail concepts are comparably represented.
Noisy alt-text is augmented with detailed MLLM-generated captions, and both long and short captions are retained per image to balance retrieval quality against zero-shot classification.

\noindent\textbf{Video-text.}\label{sec:data_video}
Mix42M (42M pairs) integrates multi-source public and proprietary corpora; videos are shot-segmented, filtered by quality and aesthetic scoring models, classified into $\sim$100 content categories with Qwen3-VL for balanced sampling, and recaptioned with category-specific prompts followed by quality assessment.

\noindent\textbf{Audio-text.}\label{sec:data_audio}
About 42M pairs are built on WavCaps (FreeSound, BBC Sound Effects, SoundBible, AudioSet) plus supplementary sources, balanced across speech, music, environmental, and sound-effect categories.
Audio is VAD-segmented, quality-filtered (silence, noise, synthetic audio), labeled by a sound event detection model, and recaptioned by fusing ASR transcripts, speaker diarization, and Qwen3-Omni descriptions with Qwen3; all waveforms are resampled to 16\,kHz and converted to 128-bin log-mel spectrograms.

\noindent\textbf{Audio-visual.}\label{sec:data_av}
About 30M temporally aligned pairs are drawn from YouTube, AudioSet, VGGSound, and YouCook2, with both streams processed by the above pipelines and cross-checked for audio-visual content consistency; independently generated audio and video captions are then fused by Qwen3, with a consistency re-annotation pass triggered upon semantic conflicts, producing semantically unified audio-visual-text triplets.


%% file: sec/5_experiments.tex
\section{Experiments}
\label{sec:exp}

\subsection{Experimental Setup}
\label{sec:exp_setup}

\noindent\textbf{Training.}
All models are initialized from the SigLIP-2 SO400M backbone~\cite{tschannen2025siglip2} and trained jointly on the four-modality data recipe (Sec.~\ref{sec:data}) with alternating-modality training and cross-modality gradient accumulation (Sec.~\ref{sec:training_strategy}); all configurations follow Table~\ref{tab:impl}.
All training and evaluation are conducted on AMD Instinct MI308X accelerators (192\,GB HBM3 per GPU) running Linux, with PyTorch~2.8, Transformers~4.57, DeepSpeed~0.17 (ZeRO-1), and FlashAttention-2 in bf16 precision.
We report \textbf{Omni-Encoder} (SO400M).

\vspace{0.3em}
\noindent\textbf{Evaluation.}
A \emph{single} shared checkpoint is evaluated zero-shot on the full benchmark suite: image classification on ImageNet-1K/A/V2/Sketch; image-text retrieval on MSCOCO; video-text retrieval on DiDeMo, MSR-VTT, ActivityNet, YouCook2, VALOR, and VATEX; audio-text retrieval on AudioCaps~\cite{kim2019audiocaps} and Clotho; and audio-visual retrieval on VALOR and VATEX.
We report classification accuracy and retrieval R@1 throughout.
Baselines are grouped into four tiers by modality coverage: image-text~\cite{radford2021learning,tschannen2025siglip2}, video-text~\cite{wang2022internvideo,li2023umt,zhao2024videoprism}, audio-text~\cite{wu2023clap,elizalde2023clap}, and omni-modal models~\cite{girdhar2023imagebind,zhu2024languagebind,bao2024onepeace,comanici2025gemini}.
``--'' denotes results that are unreported or not applicable.

\subsection{Main Results}
\label{sec:main_results}

\noindent\textbf{Image classification and retrieval.}
Despite sharing parameters with three other modality pairs, Omni-Encoder remains on par with its image-only initialization: 83.08\% on ImageNet-1K (vs.\ 84.10\% for SigLIP-SO400M), 84.12\% on ImageNet-A (+4.8 over SigLIP-SO400M), and 74.26\% on MSCOCO I$\to$T retrieval (+2.6 over SigLIP-SO400M).
Against existing omni-modal encoders, it achieves the best accuracy on all four classification benchmarks, surpassing the strongest baseline (Gemini-Omni-Embedding) by 1.6 points on ImageNet-1K.

\vspace{0.3em}
\noindent\textbf{Video-text retrieval.}
Omni-Encoder surpasses all video-specialist models on MSR-VTT (54.20\% vs.\ 51.40\% for VideoPrism-B) and ActivityNet (53.50\% vs.\ 49.60\%), and remains competitive on VATEX (55.10\%).
It consistently outperforms prior omni-modal encoders such as ImageBind, LanguageBind, and ONE-PEACE on all six benchmarks, and beats MiCo on five of six.
VLM-derived embedders with several-fold larger capacity (e.g., Qwen3-VL-Embedding-2B) still lead on several video benchmarks, suggesting headroom from scaling rather than from the unified design itself.

\vspace{0.3em}
\noindent\textbf{Audio-text retrieval.}
For audio-text retrieval we additionally apply a training-free score calibration at inference: each query's similarity logits are scaled by a suitably chosen temperature $\tau$ and reweighted by their softmax over the candidate set, $\tilde{s}=\tau s\cdot\mathrm{softmax}(\tau s)$, which sharpens the margin between top-ranked and remaining candidates without altering the learned embeddings.
Since audio-text logits occupy a different scale from vision-language ones, this per-query normalization adaptively calibrates the score range; it is applied only to audio retrieval, while all other tasks are evaluated with raw cosine scores under the standard protocol.
With 42.20\% R@1 on AudioCaps, Omni-Encoder slightly surpasses the best audio-specialist model (GLAP, 41.70\%) and matches it on Clotho (19.40\%), while exceeding all prior omni-modal encoders by at least 18 points (best: ONE-PEACE, 23.84\%).
This shows that the audio modality is not compromised by sharing the backbone with visual modalities; if anything, it benefits from the joint recipe.

\vspace{0.3em}
\noindent\textbf{Audio-visual retrieval.}
Omni-Encoder attains the best VATEX AV$\to$T retrieval (72.40\%, +11.8 over Gemini-Omni-Embedding) and the second-best VALOR T$\to$AV (68.50\%).
MiCo leads on VALOR T$\to$AV (73.38\%) but sacrifices unimodal capability (54.12\% on ImageNet-1K, 12.10\% on AudioCaps), whereas Omni-Encoder maintains strong performance on all four modality pairs simultaneously.

\vspace{0.3em}
\noindent\textbf{Cross-modal transfer.}
Pooling the evidence across tasks suggests that joint training redistributes capability rather than merely diluting it.
Relative to the SigLIP-SO400M initialization, Omni-Encoder gains on ImageNet-A (+4.8) and MSCOCO I$\to$T (+2.6); its audio embedding surpasses the best audio specialist (42.20\% vs.\ 41.70\% on AudioCaps) despite sharing the backbone with two visual modalities, and the largest margin appears precisely where cross-modal alignment is tested---VATEX AV$\to$T improves over the best prior omni-modal result by 11.8 points.
The main trade-offs are concentrated on style-shifted classification (IN-V2 $-1.7$ and Sketch $-3.5$ relative to the initialization) and on temporally demanding video retrieval (DiDeMo, YouCook2), where larger VLM-derived embedders still lead.
We attribute the gains to the temporally aligned audio-visual supervision, which provides cross-modal signals unavailable to any single-modality recipe, and the remaining gaps to capacity rather than interference.

\vspace{0.3em}
\noindent\textbf{Summary.}
One jointly trained checkpoint matches or surpasses per-modality specialists on image classification, MSR-VTT/ActivityNet video retrieval, and AudioCaps audio retrieval, and stays competitive on the remaining benchmarks.
These results support our central finding that the four modalities can mutually benefit one another under a shared backbone.
To our knowledge, this is among the first demonstrations that a deliberately simple, end-to-end shared-backbone recipe can remain competitive with per-modality specialists, and we hope it inspires further work toward even simpler omni-modal encoders.

%% file: sec/6_conclusion.tex
\section{Conclusion}
\label{sec:conclusion}

We have presented Omni-Encoder, a general multimodal encoder that unifies image, video, audio, and audio-visual understanding within a single SigLIP-2 Vision Transformer with fully shared parameters.
The design combines modality-native tokenizers, a multimodal rotary position embedding parameterized by absolute timestamps, segment-level audio-visual interleaving, and alternating-modality training with cross-modality gradient accumulation, supported by a balanced four-modality data recipe.
Across a broad benchmark suite covering classification and retrieval in all four modality pairs, a single jointly trained checkpoint matches or surpasses per-modality specialist models, while prior unified encoders still lag behind dedicated ones---demonstrating that the four modalities can be mutually beneficial under a shared backbone rather than competitive.
We hope this work offers a practical step toward general-purpose multimodal representations, where one encoder replaces a collection of per-modality models.


\noindent\textbf{Limitations.} We discuss two limitations that point to actionable future directions.
First, \emph{inference efficiency}: the shared backbone is trained to serve all modalities at once, so single-modality deployment currently carries the full model capacity; per-modality distillation or structured pruning from the shared checkpoint is a natural next step to reduce serving cost without retraining.
Second, \emph{per-modality performance alignment}: on a few specialized tasks (e.g., long-form speech understanding and fine-grained audio event detection), dedicated single-modality models trained on larger in-domain corpora still hold an edge; we expect continued scaling of the audio and audio-visual data recipe to further close this gap, and our alternating-modality training framework directly accommodates such data scaling.
Neither limitation affects the central finding that one shared parameter set can serve all four modality groups competitively.
